# Temporally Object-based Video Co-Segmentation

Michael Ying Yang[1]*, Matthias Reso[2]*,
Jun Tang[2], Wentong Liao[2], and Bodo Rosenhahn[2]

[1] Scene Understanding Group, University of Twente
[2] Institute for Information Processing, Leibniz University Hannover

**Abstract.** In this paper, we propose an unsupervised video object co-segmentation framework based on the primary object proposals to extract the common foreground object(s) from a given video set. In addition to the objectness attributes and motion coherence our framework exploits the temporal consistency of the object-like regions between adjacent frames to enrich the set of original object proposals. We call the enriched proposal sets temporal proposal streams, as they are composed of the most similar proposals from each frame augmented with predicted proposals using temporally consistent superpixel information. The temporal proposal streams represent all the possible region tubes of the objects. Therefore, we formulate a graphical model to select a proposal stream for each object in which the pairwise potentials consist of the appearance dissimilarity between different streams in the same video and also the similarity between the streams in different videos. This model is suitable for single (multiple) foreground objects in two (more) videos, which can be solved by any existing energy minimization method. We evaluate our proposed framework by comparing it to other video co-segmentation algorithms. Our method achieves improved performance on state-of-the-art benchmark datasets.

## 1 Introduction

Video object segmentation aims to group the pixels over frames of a video into spatial-temporal coherent regions, i.e., to find those pixels belonging to the same foreground object(s) in each frame. Most of the algorithms for video object segmentation focus on single object scenarios in one video sequence, e.g. [1,2,3].

However, in practical scenarios, the video contents are much more complicated and diverse. For instance, most videos contain more than one object; some foreground objects arise in indistinguishable motion or are surrounded by the background which is similar in appearance to the objects. In such circumstances, using the joint information from other videos containing the same objects, can help us to discover the foreground objects much more precisely. This method is known as *video co-segmentation*, firstly introduced by Rubio et al. in [4], which segments the common regions appearing over all the frames of two or more given video sequences containing the same objects.

**\*The first two authors contribute equally to this paper.**

　　While the results look promising, the task to distinguish and extract the foreground objects by using only the joint appearance among the videos still remains unsolved. In this paper, we propose a general framework for video co-segmentation based on the object proposals, which is a graphical model for single or multiple foreground objects in two or more videos. Our algorithm differs significantly from others mainly in two parts:

- we refine and expand the original set of object proposals from [5] by predicting them onto adjacent frames using temporal coherence information to create temporal proposal streams as well as
- formulate the selecting problem of the object proposal streams as a portable conditional random field (CRF) model.

Our first contribution exploits that the appearance and shape of objects are assumed to vary slowly over frames. We use the information of temporally consistent superpixels to create temporal proposal streams which represent a temporally consistent object-like region in the video. By the second contribution multiple foreground objects can be dealt with more easily. Our model can be solved by any existing energy minimization method. We validate our framework using two public benchmark datasets, MOViCS [6] and ObMiC [7], and compare our results with the state-of-the-art methods. An overview of our framework is illustrated in Fig. 1.

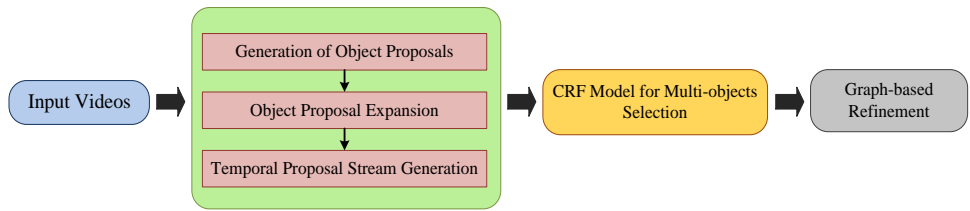

**Fig. 1.** The overview of our proposed framework. Firstly, given a video set, a group of primary object proposals for each video frame is generated. Next, we expand the original proposal set with the predicted ones based on temporal information from adjacent frames, and then combine the similar proposals from each frame in a video as the temporal proposal streams. Then the CRF model selects the best proposal stream for each object in each video. Finally each segments of each stream is refined by a spatial-temporal graph model.

## 2   Related Work

*Image Co-Segmentation* The concept of co-segmentation was firstly applied to image pairs by Rother et al. in [8]. They proposed a method to discover the common object by considering the joint information from a pair of images. Vicente et al. [9] extended the method to image based object co-segmentation by using the object proposal from [5] to segment similar objects from image pairs. Moreover, the methods proposed by [10,11] dealt with multiple object classes using discriminative clustering.

*Video Object Segmentation* Many research has been done on video object segmentation, i.e. to separate the objects from the background in videos. In contrast to the methods based on low-level features, several approaches adopted the 'objectness' measure to seek for the object-like proposals for primary video object segmentation in a single video [1,2,12]. In addition, Grundmann et al. [13] clustered a video into spatio-temporal consistent supervoxels, and Jain and Grauman [14] used the consistency of these supervoxels as a higher order potential for semi-supervised foreground segmentation.

*Video Object Co-segmentation* Recently, an increasing number of methods focus on video object co-segmentation. The method proposed in [4] grouped the pixels into two levels: the higher level consists of space-time tubes, and the lower one is composed of region segments within the frames. Based on initial foreground and background estimation and dense feature extraction of the regions and tubes, they constructed a probabilistic model of the foreground and background, and iteratively refined the results and updated the model. The supervoxels based method proposed in [15] employed dense optical flow to derive the intra-video relative motion and Gaussian mixture models to characterize the common object appearance. Both methods can only deal with the videos containing a single common object. In [6], Chiu and Fritz proposed a multi-class video object cosegmentation method using a non-parametric Bayesian model to learn a global appearance model, which connected all the segments of the same object. However, it is based on low-level descriptors for grouping the foreground pixels into classes. Fu et al. [7] built a standard multi-state selection graph model (MSG) in view of the intra- and inter-video coherence. Guo et al. [16] considered also the persistence of different parts of the foreground during the video and also proposed automatic model selection while binding them together. In all of these methods, [7] achieves the best segmentation results. However, the MSG certainly assigns each node (frame) an optimal label (object proposal), which means they can not find the objects when it does not appear in the first frames. Besides, if the object is totally covered in some frames, there would be no selectable proposal to represent the object, even so the MSG still chooses one for these frames, which fulfills the lowest MRF energy. Although they used a graph-based segmentation for refinement, it is still unrecoverable in case of the wrong proposals characterized by low-level features which are similar to the object of interest. Furthermore, their graph model is constructed with fully-connected states of each node between the multiple videos, which costs lots of time in comparison between different states.

Our proposed method based on temporal proposal streams overcomes aforementioned challenges. All the streams are generated by the detected similar proposals from each frame without the limitation of starting or ending point. Similar streams are merged into a single stream via spectral clustering, even if they are not completely consecutive. Moreover, our framework is more efficient to obtain the final results, due to the fewer comparisons between the states (temporal proposal streams).

## 3    Proposed Method

Given a set of $N$ videos as $\{V^1, \ldots, V^N\}$ we primarily achieve a group of object-based proposals using [5] in each frame $f_t^n$, $n \in N$, $t \in F_n$. These proposals $p_t^n$ are generated by performing graph cuts based on a seed region and a learned affinity function. They are also scored from best to worst based on a ranking system. These candidates are used as input of our proposed video co-segmentation method.

### 3.1    Object Proposal Expansion

In order to find the object-like candidates among them, we define a score as proposed by [1] based on appearance cues and salient motion patterns relative to their surroundings:

$$A(p_{t_i}^n) = O(p_{t_i}^n) + M(p_{t_i}^n), \tag{1}$$

where the score $A(p_{t_i}^n)$ of $i^{th}$ proposal in frame $f_t^n$ is constituted by the static intra-frame objectness score $O(p_{t_i}^n)$ and the dynamic inter-frame motion score $M(p_{t_i}^n)$. The objectness score $O(p_{t_i}^n)$ is the original score in the proposal-generating process from [5]. It reflects how likely the proposal $p_{t_i}^n$ is a whole object. The motion score $M(p_{t_i}^n)$, as defined in (2), measures the confidence that the proposal $p_{t_i}^n$ corresponds to a coherently moving object in the video.

$$M(p_{t_i}^n) = 1 - exp(-\frac{1}{\bar{M}}\chi^2_{flow}(p_{t_i}^n, \overline{p_{t_i}^n})), \tag{2}$$

where $\overline{p_{t_i}^n}$ denotes the pixels around the proposal $p_{t_i}^n$ within a loosely fit bounding box, and $\chi^2_{flow}(p_{t_i}^n, \overline{p_{t_i}^n})$ is the $\chi^2$-distance between $L_1$-normalized optical flow histograms with $\bar{M}$ denoting the mean of the $\chi^2$-distance.

In [5] only the local information of each individual frames is considered, thereby neglecting the temporal information. Taking this into consideration, we adopt the idea of [17] to create temporally consistent superpixels (TCS) to map all the proposals onto the adjacent frames. In consequence, the TCS labels of each proposal may guide us to predict an additional successive proposal in these frame.

As illustrated in Fig. 2, each proposal of frame $f_t^n$ is warped by selecting the superpixels with the same TCS labels on frame $f_{t+1}^n$. Therefore, the new predicted proposal contains the TCS labels from $p_{t_i}^n$. We refine it using graph-based image segmentation [18]. With this predicted proposal, we seek for a proposal in frame $f_{t+1}^n$ which is similar to the newly predicted one. The intersection-over-union overlap ratio is defined as the judgment criteria as follows:

$$o = \frac{|Warp_{t \to t+1}(p_{t_i}^n) \cap p_{t+1_j}^n|}{|Warp_{t \to t+1}(p_{t_i}^n) \cup p_{t+1_j}^n|}. \quad (3)$$

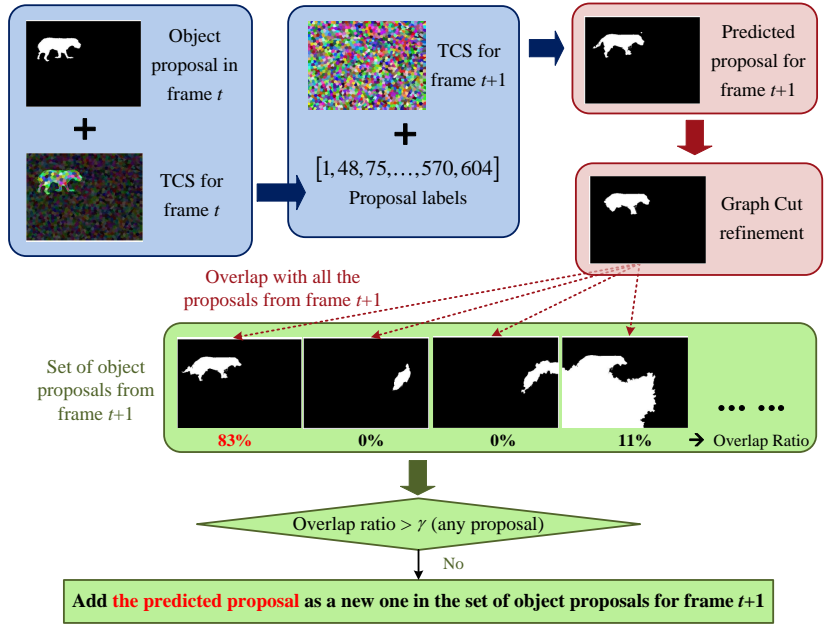

**Fig. 2.** Object proposal expansion procedure. For each object proposal in frame $t$, we predict its temporally consistent one for frame $t+1$ based on the TCS labels and overlap it with all the original existing proposals. If none of them has the overlap ratio higher than the threshold, we add the predicted proposal as an additional one in the proposal set of frame $t+1$.

If any proposal $p_{t+1_j}^n$ in frame $f_{t+1}^n$ does not have an overlap ratio larger than the threshold $\gamma$ ($\gamma$ is set to 0.7 in this paper), we use the predicted $Warp_{t \to t+1}(p_{t_i}^n)$ as an additional proposal and add it to the proposal set of frame $f_{t+1}^n$. In practice, this procedure is carried out in a consecutive fashion in both forwards and backwards direction. This ensures that any missing proposal if properly propagated onto every frame.

### 3.2 Temporal Proposal Streams

Based on the expanded proposals, we discover the groups of temporal proposal streams which may represent a consistent foreground object-like region in the video.

Primarily, we start to generate the streams for video $V^n$ from its first frame $f_1^n$. The $x$ most highly ranked proposals in the first frame are assigned as the beginning of the $x$ initial proposal streams. Then we seek a similar proposal in the next frame for each of the stream with the overlap of the TCS labels, as mentioned in Sec. 3.1. The one with the highest overlap ratio will be regarded as a new member in the corresponding proposal stream. If a proper proposal can not be found in this frame, this stream ends up here; otherwise, the process moves on to next frame. Meanwhile, we also consider the $x$ most highly ranked proposals in the following frames $f_k^n$. Some of them may be already connected with the existing streams, and the rest are used to start new streams. So, in practice, the set of the streams grows over frames. But with the limitation of $x$, it will not grow too much, because most of the highly ranked proposals of each frame should just continue the already started streams. On the other hand, this growing process helps us to find new objects which maybe does not show up in the first frame.

In some cases, the object is totally occluded in some frames and then shows up again, as shown in Fig. 3. Our aforementioned method treat it as two different streams which are supposed to represent the same object. To solve this problem, we need to bond some of the generated streams. Before the combination, the streams which span all the frames are retained unchanged, while the ones containing only one frame are abandoned. For the rest of them, we adopt the spectral clustering based on their colour appearance to group them in $y$ clusters.

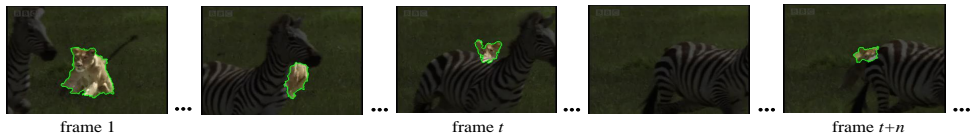

**Fig. 3.** Object occlusion occurring in a video of 'LionsAll' video set in MOViCS dataset [11]. We retain the proposal before the occlusion and use its appearance to compare with the proposals when the lion shows up again.

### 3.3  CRF Model for Multi-object Selection

Since the graphical model provides a standard framework for capturing complex dependencies among random variables, it helps us to select the most object-like temporal proposal stream for each video as the object segmentation. In this paper, this problem is formulated as a graphical model in the form of a conditional random field (CRF), as illustrated in Fig. 4.

Each node represents an object in a video and the possible states are comprised of the corresponding temporal proposal streams. We seek a proper stream

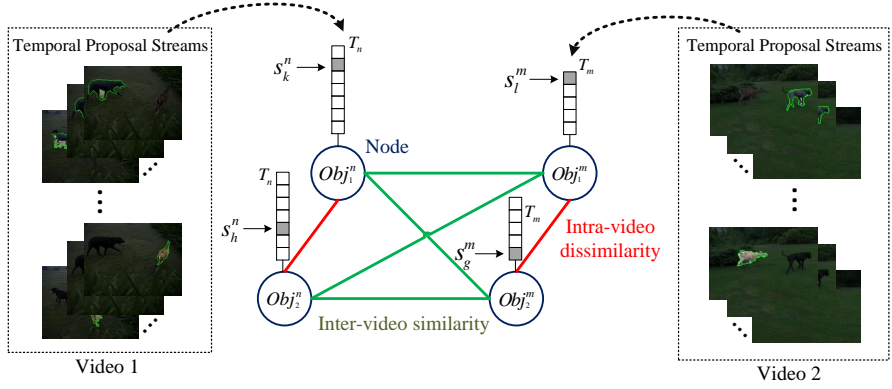

**Fig. 4.** Our multi-object selection graphical model for the 'Dog and Deer' video set from ObMiC dataset [7], which contains two video sequences with two objects.

to represent the object for each node. The energy function of the graphical model is defined as:

$$E = \sum_{n=1}^{N}\sum_{k=1}^{C_n} E_{unary}(s_k^n) + \alpha_1 \cdot \sum_{n=1}^{N}\sum_{\substack{k,h=1\\k\neq h}}^{C_n} E_{intra}(s_k^n, s_h^n) + \alpha_2 \cdot \sum_{\substack{n,m=1\\n\neq m}}^{N}\sum_{k=1}^{C_n}\sum_{l=1}^{C_m} E_{inter}(s_k^n, s_l^m), \quad (4)$$

where $\alpha_1$ and $\alpha_2$ are weighting coefficients.

The unary energy uses the aforementioned score $A(s_k^n)$ from Sec. 3.1 and the saliency score $S(s_k^n)$ of all the regions in each stream to represent its likelihood belonging to the foreground:

$$E_{unary}(s_k^n) = -log[max(\bar{A}(s_k^n), S(s_k^n))], \quad (5)$$

where $\bar{A}(s_k^n)$ is the mean score of all the proposals in stream $s_k^n$. Due to the irregular movements of the foreground objects, we also consider their saliency as a supplementing static cue. For the whole video, we compute the co-saliency map based on all the frames using [19] and get the saliency score from the overlap between the region and the corresponding map.

The pairwise term includes two parts, intra- and inter-video energy. $E_{intra}(s_k^n, s_h^n)$ is the intra-video energy between the streams $s_k^n$ and $s_h^n$ in video $V^n$, which represents the stream similarity penalty. Each object has its own stream in a video, which means a stream should not be assigned to different objects. Thus, the stream similarity penalty is described by the dissimilarity between the streams:

$$E_{intra}(s_k^n, s_h^n) = -log(D_f(s_k^n, s_h^n)), \quad (6)$$

where $D_f$ is the low-level feature similarity between them, which is defined as:

$$D_f(s_k^n, s_h^n) = \frac{1}{M_m}\chi_f^2(s_k^n, s_h^n). \qquad (7)$$

In practice, we compare all the regions of both streams and average the scores. $\chi_f^2(s_k^n, s_h^n)$ is the weighted combination of the $\chi^2$-distances between the normalized colour histograms and the shape histograms of $s_k^n$ and $s_h^n$. $M_m$ denotes the mean value of the $\chi^2$-distance. In our work, shape is represented by the HOG descriptor [20] within a minimum bounding box enclosing the region.

The other pairwise term $E_{inter}(s_k^n, s_l^m)$ measures the object consistency among different videos. In this graph, each stream from one video is connected to those in the other videos. We define the inter-video energy as:

$$E_{inter}(s_k^n, s_l^m) = D_f(s_k^n, s_l^m), \qquad (8)$$

in which $D_f$ is the low-level feature similarity computed by (7).

For inference, we employ TRW-S [21] to find the approximated labelling that minimizes the energy function. Since the original object proposals generated by [5] are only roughly segmented, we refine the final results as [1] with a pixel-level spatio-temporal graph-based segmentation to achieve a better segmentation.

## 4   Experiments

We implement our proposed method in MATLAB and compare it against four state-of-the-art methods related to video co-segmentation: Multi-class video co-segmentation (MVC) [6], Object-based multiple foreground video co-segmentation (ObMiC) [7], Extracting primary objects by video co-segmentation (EPOVC) [22] and the latest Consistent foreground co-segmentation (CFC) [16]. For the comparison we use two state-of-the-art datasets: Multi-Object Video Co-segmentation (MOViCS) dataset [6] for single object video co-segmentation and Object-based Multiple Foreground Video Co-segmentation (ObMiC) dataset [7] for the multiple objects case. Same as in [6], the *intersection-over-union metric* (IOU), defined as $\frac{R \cap GT}{R \cup GT}$, is used as evaluation metric in this paper.

*Implementation Details* For both datasets, the number of TCS in each frame of each video sequence is around 1500, which makes sure that each TCS represents a region with a proper size containing consistent appearance. The threshold in the propagation procedure of object proposals is defined as 0.6, which judges whether a new additional proposal would be added in the original proposal set of next frame. When we discover the temporal proposal streams for each video, we use the 40 most highly ranked proposals in the first frame to initialize the streams as their beginning. In addition, 10 most highly ranked proposals in the following frames are considered as the candidates to start a new stream. After generating the streams, all the incomplete streams are grouped into 20 or 5 clusters, which depends on the amount of the incomplete streams. All the low-level features leveraged in the framework consist of colour information and shape information. The colour feature is computed in Lab colour space and RGB colour space with 117 bins and the shape information from HOG descriptor is presented in 81 bins. To combine these two features, we set the weighting coefficient for the colour as 2 to increase the weight. As for the graphical model, the weighting coefficients $\alpha_1$ and $\alpha_2$ to balance the two pairwise potential terms is set empirically.

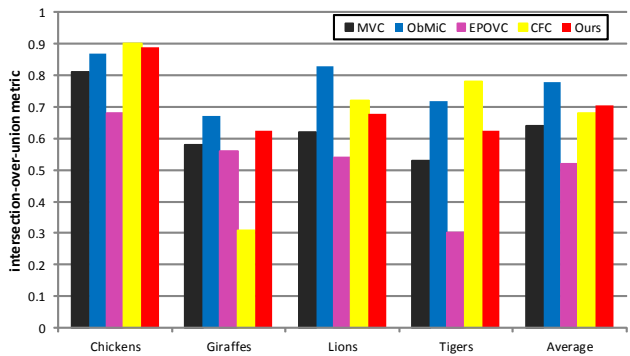

**Fig. 5.** The IOU metric on MOViCS dataset [6].

*Evaluation on MOViCS Dataset* We test our framework on the MOViCS dataset [6], which includes four different video sets and 11 videos in total. Each video set contains one or two objects, and for five frames of each video a ground truth labelling is provided. All objects appear in the videos of this dataset in an irregular way. Although some videos comprise more than one foreground object, we only consider the object appearing in each video of the video set.

As shown in Fig. 5, our proposed framework outperforms the multi-class video co-segmentation method of [6] significantly. Using the temporal coherence between the adjacent frames improves the segmentation results. Comparing with the ObMiC from [7], we have better results in one video set. The reason for the difference in the other video sets is that they employed all object proposals in each frame as candidates for their graphical model, which chooses the proper proposal for each frame separately. This low-level method keeps more details for each proposal, but loses some temporal relevance between the proposals. Besides, the computational overhead for the fully connected graphical model is much higher as more similarities have to be evaluated. In comparison to EPOVC [22] which has a similar structure as ObMiC our method produces a higher average accuracy. They applied only the low-level feature of each proposal to build the graphical model, which is restricted by its initial configuration. Although the recently published CFC method of [16] automatically chooses a suitable model for each video set and performs well on the 'Tigers' sequence we achieve on par

10   Yang et al.

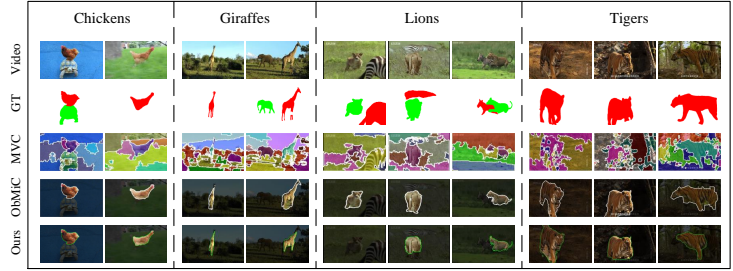

**Fig. 6.** Single object co-segmentation results on MOViCS dataset. First row is the sample frames of given videos; second row represents the ground truth; from the third to fifth row are the segmentation results from MVC [6], ObMiC [7] and our framework, respectively.

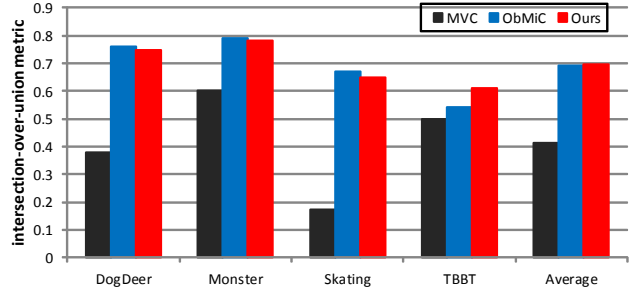

**Fig. 7.** The IOU metric on ObMiC dataset [7].

or better accuracy on the other three video sets. Figure 6 shows some qualitative segmentation examples.

*Evaluation on ObMiC Dataset* The ObMiC datset [7] comprises four video pairs, each containing two common foreground objects. The scenarios of these video sets are completely different and a ground truth labelling is provided for all frames.

In the first three video sets our accuracy is better than MVC but lower than ObMiC as shown in Fig. 7. But in the last video set, our accuracy is superior to theirs.

The segmentation results of **DogDeer** in the fist column in Fig. 8 show that ObMiC segments the boundaries of objects slightly better than us, which are similar to the results from the second video set **Monster** in the second column. A more complicated environment is about the reality scene with human beings. In the third column, our segmentation results in sequence **Skating** are better than MVC and ObMiC. MVC segments the bodies in many pieces and ObMiC can only find partial region of the objects, in which the colour appearance is consistent. In the last **TBBT** video set, our framework outperforms the other two methods in intersection-over-union metric. The segmentation results from ObMiC also has the problem in the last video set, that they only find the clothes. From this aspect, our results are better for the woman, but yet to be satisfied for the man.

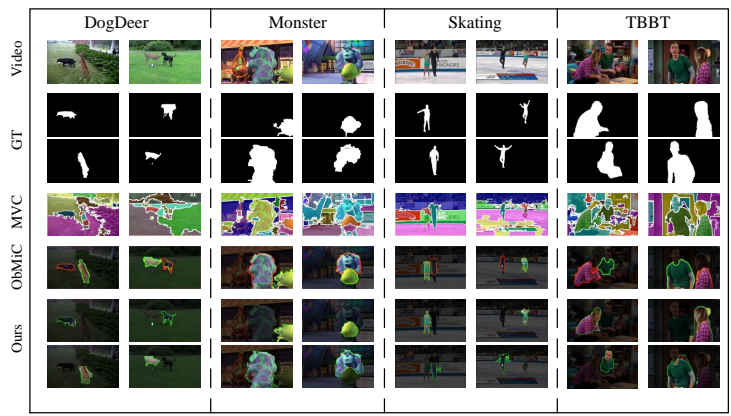

**Fig. 8.** Multiple objects co-segmentation results on ObMiC dataset. First row is the sample frames of given videos; second row represents the ground truth; from the third to fifth row are the segmentation results from MVC [6], ObMiC [7] and our framework, respectively.

## 5  Conclusion

We propose a video co-segmentation framework to extract the common foreground object(s) from the given video set. The procedure consists of two key steps: based on the basic object proposals, we firstly use the temporal information between the frames to combine the consistent proposals together as temporal proposal streams; secondly, a stream for each object is selected in each video by the CRF model depending on their appearance. Our framework is not restricted in the number of objects or videos, and it outperforms most of the state-of-the-art methods in terms of accuracy with a lower computational burden on both state-of-the-art benchmark datasets for the video object co-segmentation task.

## Acknowledgements

The work is partially funded by DFG (German Research Foundation) YA 351/2-1. The authors gratefully acknowledge the support.

12      Yang et al.